\documentclass[conference]{IEEEtran}
\IEEEoverridecommandlockouts
\usepackage{cite}
\usepackage{amsmath,amssymb,amsfonts}
\usepackage{algorithmic}
\usepackage{graphicx}
\usepackage{textcomp}
\usepackage{booktabs}

\def\BibTeX{{\rm B\kern-.05em{\sc i\kern-.025em b}\kern-.08em
    T\kern-.1667em\lower.7ex\hbox{E}\kern-.125emX}}
\begin{document}

\title{A deep learning approach for understanding natural language commands for mobile service robots}

\author{\IEEEauthorblockN{ Pedro Henrique Martins}
\IEEEauthorblockA{
\textit{Instituto Superior T\'{e}cnico}\\
Univ Lisboa\\pedrohenriqueamartins@gmail.com}
\and
\IEEEauthorblockN{ Lu\'{i}s Cust\'{o}dio}
\IEEEauthorblockA{\textit{Institute for Systems and Robotics} \\
\textit{Instituto Superior T\'{e}cnico}\\
Univ Lisboa\\
luis.custodio@isr.tecnico.ulisboa.pt}
\and
\IEEEauthorblockN{Rodrigo Ventura}
\IEEEauthorblockA{\textit{Institute for Systems and Robotics} \\
\textit{Instituto Superior T\'{e}cnico}\\
Univ Lisboa\\
rodrigo.ventura@isr.tecnico.ulisboa.pt}
}
\maketitle

\begin{abstract}
Using natural language to give instructions to robots is challenging, since natural language understanding is still largely an open problem. In this paper we address this problem by restricting our attention to commands modeled as one action, plus arguments (also known as slots). For action detection (also called intent detection) and slot filling various architectures of Recurrent Neural Networks and Long Short Term Memory (LSTM) networks were evaluated, having LSTMs achieved a superior accuracy. As the action requested may not fall within the robot’s capabilities, a Support Vector Machine(SVM) is used to determine whether it is or not. For the input of the neural networks, several word embedding algorithms were compared. Finally, to implement the system in a robot, a ROS package is created using a SMACH state machine. The proposed system is then evaluated both using well-known datasets and benchmarks in the context of domestic service robots. 
\end{abstract}


\section{Introduction}

Robotics holds tremendous potential to benefit humans in every aspect of life. These benefits have been increasingly visible in industrial environments, such as factories. Due to the technological evolution, robots are also starting to be integrated into the human environment, for everyday use. This integration is more likely to succeed if robots are able to interact with humans. Since language is the most natural way of communication among humans, is expected to play an important role in this context.

The purpose of this work is to develop a system that is able to understand the action requested by a command in natural language, being the process divided in two steps. The first, action detection corresponds to determine the action the robot has to perform, such as \textit{motion} or \textit{guiding}. The aim of the second, slot filling, is finding the command arguments, for instance \textit{location} or \textit{person}.  

This work was initially motivated by the need of an effective speech understanding functionality for the @Home robotic competitions, such as RoboCub@Home \cite{x} and European Robotics League (ERL) \cite{y}, to be used by the SocRob@Home team \cite{z}. However, we tried to develop a system that would be able to be used in real situations instead of just caring about the competition's results.

\section{Related Work}

Slot filling corresponds to the process of automatically extracting semantic concepts, filling a set of arguments or slots. Most of the first approaches to solve this were based on syntactic and semantic grammars, such as TINA \cite{tina} and Gemini \cite{gemini}. Due to the difficulty of creating grammars, discriminative  classifiers, such as Conditional  Random Field and SVM and generative classifiers, like Finite State Trandsucers and Hidden Markov Models (HMM), started being used for this task \cite{ricky}.

The task of action detection aims at classifying a speech utterance into one of various semantic classes. 
The first reported works in this area were developed for call routing, trying to direct the call to the right operator \cite{HMIHY,carpenter}. With the advances in discriminative classifiers such as, Boosting \cite{adaboost}, SVM \cite{svm} and Minimum Classification Error \cite{kuo}, researchers started using them in action detection problems.

Similarly to this work, LU4R, Language Understanding chain For Robots, is a Natural Language Understanding system that has as one of its purposes the recognition of the RoboCup GPSR instructions \cite{lu4rp,lu4ra}. In this system, a statistical semantic parser and a combination of a structured SVM and an HMM are used.

 With the increase of available data and computational power, deep learning algorithms started achieving state of the art results in a wide range of areas, including NLP.
 
Ravuri and Stolcke applied RNNs and LSTMs to the action determination problem \cite{ravuri}. For the ATIS dataset both RNNs and LSTMs had better performances than a Maximum Entropy language model also tested. However, the RNN performed slightly worse than a boosting model. For the Conversational Browser (CR) dataset RNNs and LSTMs outperformed other models, but in contrast to the ATIS dataset, LSTMs performed worse than RNNs. This contrast is due to the fact that in ATIS, sentences are bigger \cite{ravuri}, being the peak length 11 words. In the CR dataset most sentences have less than 5 words. 
  
Kaisheng Yao et al. compared RNNs with FST, SVM and CRF in the ATIS slot filling task \cite{rrnlu}. The results showed that RNNs outperform all the previous methods used for this task.

Gr\'{e}goire Mesnil et al. also tested DRNNs and DBRNNs on a slot filling task \cite{mesnil}. To test the performance and compare to CRFs, they used the ATIS dataset, a movies dataset and an entertainment dataset. For the movies dataset and ATIS, the performance of RNNs was better than CRFs. However, for the entertainment dataset the performance was slightly worse, being their explanation the fact that in the entertainment domain the slots correspond to bigger expressions such as movie names \cite{mesnil}.

Researchers, \cite{slulstm,mdjs}, have also compared the performance of RNNs and different types of LSTMs in the ATIS slot filling task, concluding that the best results are achieved when using deep and bidirectional LSTMs.

\section{Problem description}

In this paper we restrict our attention to commands given by a human to a robot. A command comprises an action, among a pre-defined set of actions the robot is capable of performing, and zero or more arguments, hereby called slots. Given a command, the robot has to determine not only the action but fill the slots.

These two requirements can be achieved by doing action detection and slot filling. Also, if the number of actions the robot is able to perform is large, these can be divided in domains, being the domains determined before. Then depending on the domain, the action is detected. For the sake of simplicity, we assume a single domain throughout this paper.

\begin{figure}[h!]
\centering
\includegraphics[width=9cm]{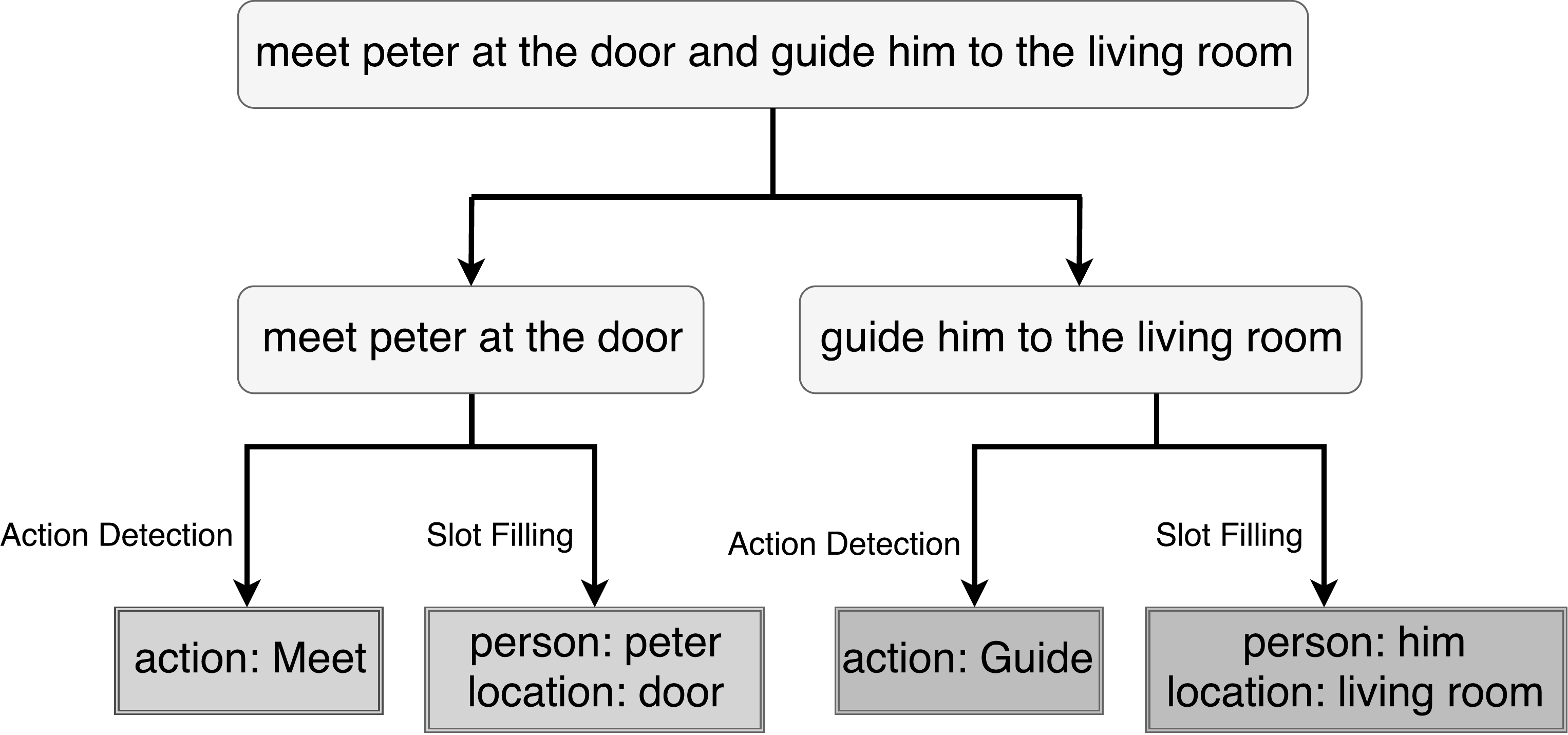}
\caption{Problem example}
\label{prob}
\end{figure}

Before applying the action detection and slot filling, as the instruction can contain various commands, the instruction sentence is divided in phrases, being each a command.
An example instruction and its breakdown is showed in figure \ref{prob}.

\subsection{The @Home Robot Competitions}

General Purpose Service Robots, GPSR\cite{gpsr}, is a task in which a voice instruction, containing one or more commands, is given to the robot, which has to understand and perform it.

Observing instructions created by the GPSR generator available publicly, a set of actions and arguments, that can be seen in table \ref{gpsrtab}, was created.

\begin{table}[h!]
  \renewcommand{\arraystretch}{} % more space between rows
  \centering
  {%
    \begin{tabular}[]{lcc}
      \toprule
        Actions &  Description \\[0.1cm]
      \midrule
        motion   &   moves to some place \\[0.05cm]
        meet   &   meets a person   \\[0.05cm]
        grasp   &   grabs a object  \\[0.05cm]
        place   & places a object  \\[0.05cm]
        take   & takes object to some place or to someone \\[0.05cm]
        tell   &  tells something to someone    \\[0.05cm]
        answer  &   waits for question   \\[0.05cm]
        find   &  looks for an object  \\[0.05cm]
        guide   &  guides a person to a location  \\[0.05cm]
        follow   &  follows a person to a location  \\[0.05cm]
      \bottomrule
    \end{tabular}
  }%
    \vspace{0.1cm}
  \caption{Set of actions and respective description for GPSR}
  \label{gpsrtab}%
\end{table}

To create the training sets, firstly the RoboCup's generator was used. However, as these generator did not make annotations, the dataset had to be small. Because of this a command generator that automatically annotates the sentences was created, allowing to create big datasets. For testing, a dataset was created using the RoboCup's generator\footnote{https://github.com/RoboCupAtHome/gpsr\_command\_generator}.

In the Speech recognition functionality benchmark, FBM3, the robot receives some audio files that has to understand and write the correspondent actions and arguments in a text file.
The dataset for FBM3 was created the same way, but differing in the set of actions and arguments, which is presented in table~\ref{parsetab}.

\begin{table}[h]
  \renewcommand{\arraystretch}{1} % more space between rows
  \centering
  {%
    \begin{tabular}[]{lcc}
      \toprule
        Actions &  Description\\[0.05cm]
      \midrule
        motion   & robot moves to some place \\[0.05cm]
        searching   &  looks for an object \\[0.05cm]
        taking   &   grabs a object    \\[0.05cm]
        placing   &   places a object  \\[0.05cm]
        bringing   & takes object to a place or to someone \\[0.05cm]
      \bottomrule
    \end{tabular}
  }%
    \vspace{0.1cm}
  \caption{Set of actions and respective description for FBM3}
  \label{parsetab}%
\end{table}

\section{System Description}
\label{sec:imple}

\begin{figure}[h]
\centering
\includegraphics[width=9cm]{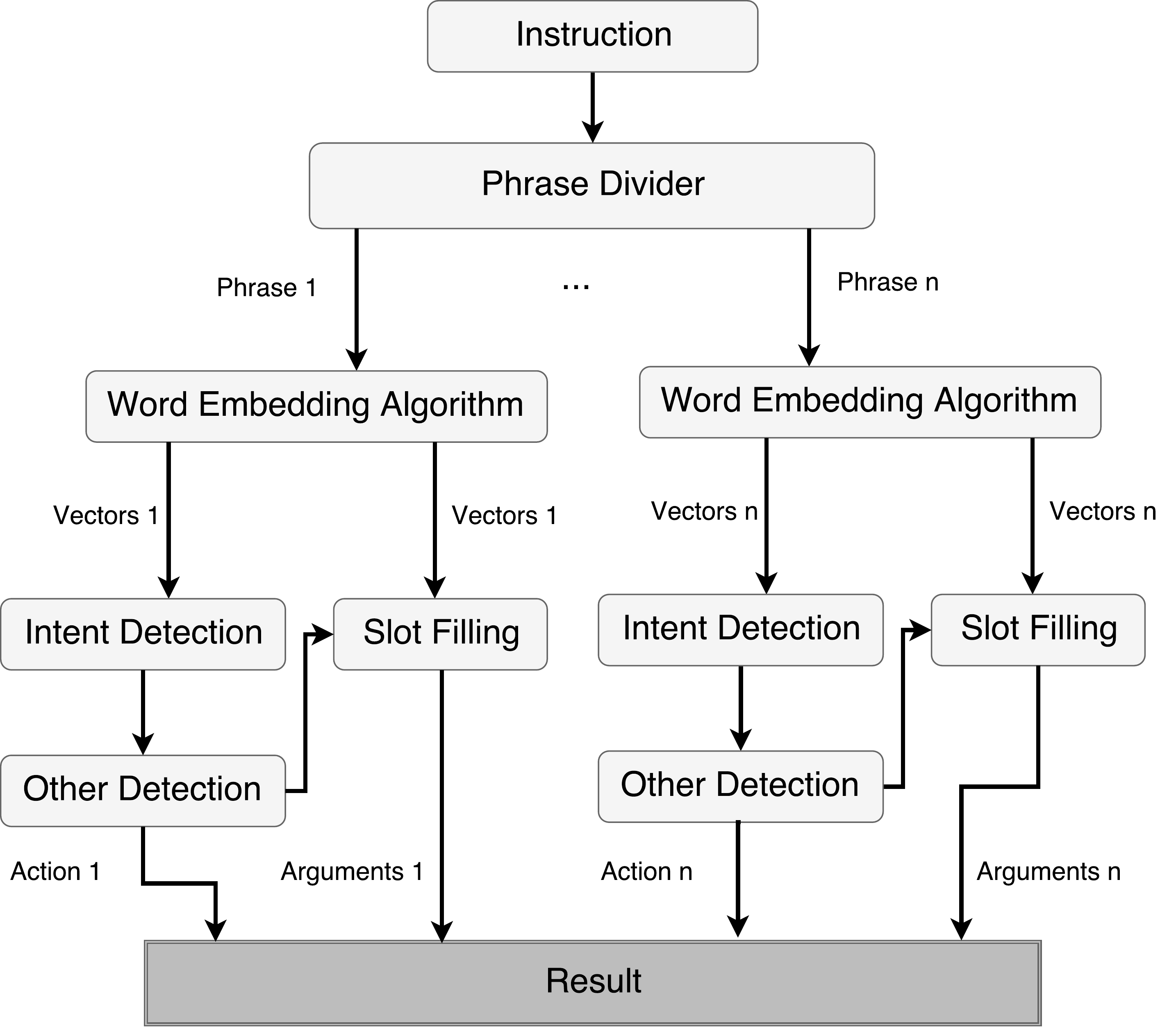}
\caption{Scheme of the first approach}
\label{scheme1}
\end{figure}

Two different approaches were implemented to understand the instructions given to the robot in natural language, being the first one shown in figure \ref{scheme1}. After the instruction is received by the robot it is recognized by the speech recognition system. The resulting \textit{string} is divided in phrases, being each resulting phrase a command to the robot. The words in the commands are then converted to feature vectors, that are the input of the action detection and slot filling networks. 
Finally, the action detection result is passed to an algorithm that determines whether the action is contained in the set actions. If it is not, the result of the slot filling classification is erased and the resultant action corresponds to \textit{Other}. Otherwise, the result is the action detected and the slots identified.   

\begin{figure}[h]
\centering
\includegraphics[width=9cm]{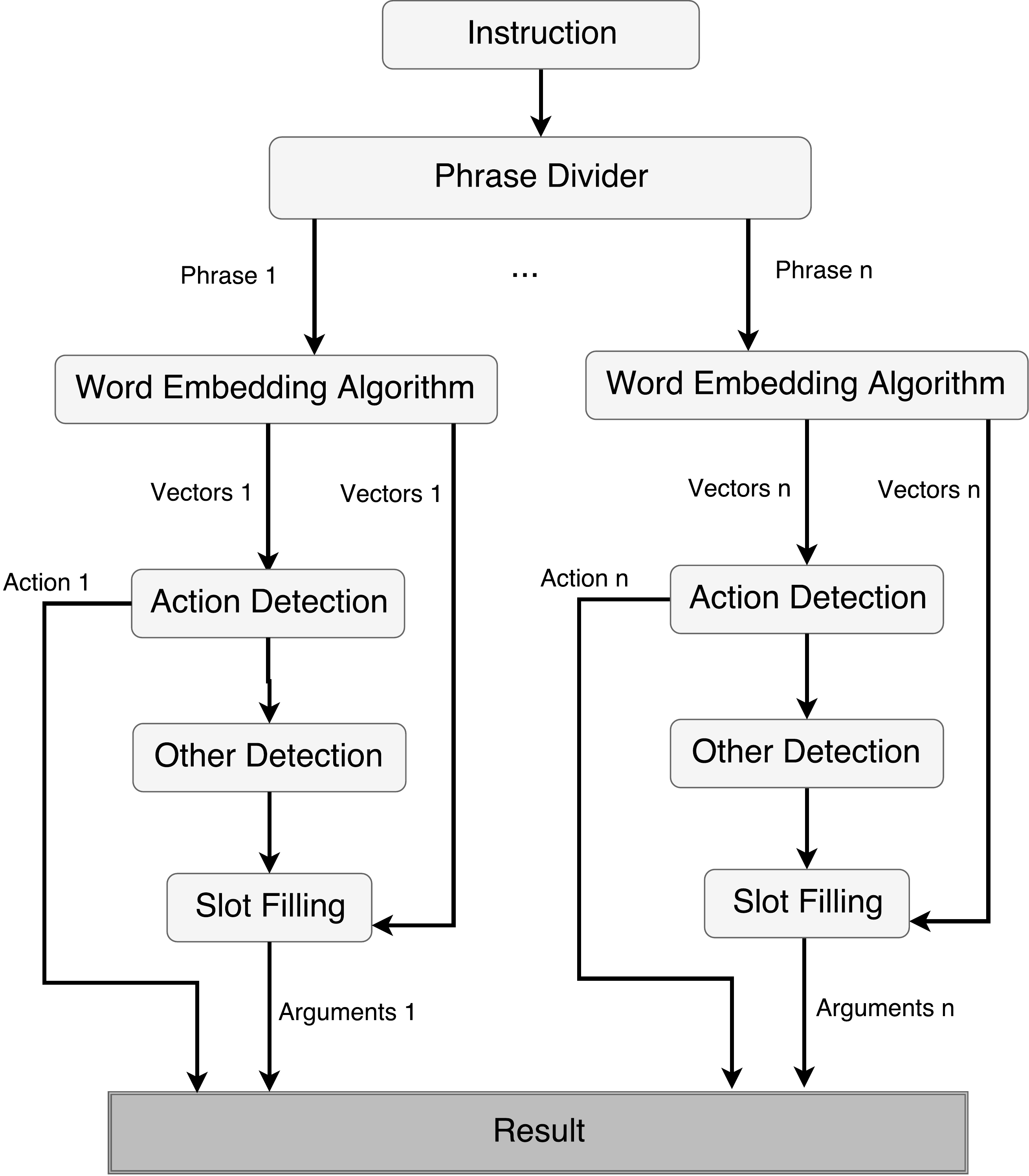}
\caption{Scheme of the second approach}
\label{scheme2}
\end{figure}

In the second approach, represented in figure \ref{scheme2}, instead of having the same slot filling model for all the actions that can be detected, there is a different model for each one. The slot filling model used is chosen accordingly to the result of the action detection result.

\subsection{Dividing instructions in phrases}

The instructions given to a service robot often have more than one command.
Because of that the instruction given has to be divided in phrases, that correspond to commands.

To achieve this an open source syntactic parser from Google was used, Parsey McParseface, which is part of Syntaxnet \cite{syntaxnet}. Given a sentence as input, Syntaxnet attributes a part-of-speech, POS, tag to each word, describing its syntactic function and determines the relationships between the words.

To divide the sentences, firstly the existent verbs are found. The auxiliary verbs are differentiated from the principal verbs by the word they are related to, being principal verbs related with the verb of the previous phrase and auxiliary verbs related to a principal verb, which is the next or previous word in the sentence.

By assembling the words that have relationships with the principal verbs, the words that are related to that words and so on, the phrases are constituted. 
As sometimes terms are related to others in previous phrases, the word order is taken into account when dividing the sentence. This is achieved by looking at the position of the principal verbs and of conjunctions when existent.

Finally the conjunctions connecting the commands are removed, since they do not add meaning to the sentence.

\subsection{Word embeddings}

As input to the neural networks, there were two options, one-hot vectors, which contain zeros and a one at the index of the correspondent word, or word embeddings, feature vectors that try to capture the meaning of the words and can be created by several algorithms.
Although, according to some literature, models using word embeddings have better performance than using one-hot vectors \cite{rrnlu}, some cases where the difference is not significant have been reported \cite{mdjs}. Because of this, both one-hot encoding and word embeddings were evaluated.

There is also not much agreement on which model has the best performance, Word2vec skip-gram model or GloVe. As a consequence, the two methods were analyzed.

\subsubsection{Word2vec}

Several extensions to the original skip-gram model were developed. In this work, the extensions used are negative sampling, being the number of words sampled $k=15$ and subsampling of frequent words with a threshold of $10^{-5}$, as advised in \cite{dmikolov}.

The corpus used to train the model was the Wikipedia of 2014, which contains $1.6$ billion tokens. Before being used the corpus was tokenized and lowercased using the NLTK Python package\footnote{http://www.nltk.org/}. The punctuation and the numbers were also removed, since, speech recognition systems do not recognize punctuation and recognize numbers in full.

A vocabulary constituted by the $50000$ most frequent English words was used. The window size chosen was of $5$ words, the center one, two before and two after. To train the neural network the Adam optimizer \cite{adam} was used to minimize a cross entropy loss function. The learning rate chosen was fixed and of $0.01$.

\subsubsection{GloVe}

To create feature vectors using the GloVe method, the same corpora and vocabulary was used. The tokenization applied was also similar to the one in Word2vec.

By observing the tests made by Pennington et al. \cite{glove}, a vector dimension of $300$ was chosen. As for the context, a symmetric window with a size of $10$ words was used. 
The maximum number of the matrix elements, $x_{max}$, used in the weighting function was $100$ and the value of $\alpha$ was $3/4$, as recommended.

The model was trained using the Adaptive Gradient optimizer, AdaGrad \cite{adagrad}, since it has achieved very good results for problems with sparse data and was also used in \cite{glove}.

\subsection{Action detection}

In the action classification task, all the words in the instruction can be useful to determine the action the robot as to perform. To have the ability of capturing the meaning and connections between all the words in the phrase, RNNs and LSTMs were tested.

The input provided to the neural networks corresponds to a word vector, computed by one-hot encoding or a word embedding algorithm, for each word in the sentence. The output is a group of confidence values, one for each of the actions in the set being the action selected the correspondent to the bigger value. 
However, the action could be one that is not represented in the set of classes.  
Hence, it is necessary to classify if the action corresponds to the class with higher confidence value or to other action that does not belong to the set.

In the first experiments, a softmax layer was used to convert this confidence values into probabilities. However, when predicting whether the action was in the set of classes, the performance decreased using the probabilities. 

To train the LSTMs, the Adam optimizer \cite{adam} with varying learning rates was used to minimize a softmax cross entropy loss function.

\subsubsection{Action \textit{Other} determination}
\label{other}

To predict if the action that the robot has been instructed to do is the one resultant from the neural network previously explained or if it is one that was not part of the set, a SVM is used. A SVM was the algorithm chosen because this corresponds to a simple binary classification problem.  

To train the SVM, a dataset containing a subset of the data used to train the neural network for the action detection task and some commands for which the output should be $Other$, is used. The input of the SVM corresponds to the maximum value that is outputed by the action detection neural network. Although not all types of sentences with an action that is not present in the actions set are represented in the dataset, the value that they will output in the action detection neural network is similar. 
This way, the SVM works similarly to a threshold, identifying the action as $Other$ if the confidence value is not reached.

\subsection{Slot filling}

For the slot filling task, RNNs and LSTMs correspond to the network architectures tested.

The input corresponds to word vectors, like in the action determination task. The output corresponds to a tag for each word that follows the IOB format \cite{iob}. The first word of a slot is tagged with a $B$ of begin, the other words inside the slot are represented with the $I$ of inside and the words that do not belong to a slot are tagged with the $O$ of outside.

In implementation method $1$, as the possible classes are the same for all actions, if for some reason a utterance term has a tag that is not part of the set of arguments correspondent to the action detected, it is replaced by $O$. As an example, for the action "robot go to the kitchen", if "kitchen" is identified as the argument $what\;to\;tell$, it would be changed to $O$.  

The training of the neural network was performed similarly to the one for the action detection task.

\section{Results}
\label{sec:resul}

\subsection{Metrics}

The metric used to validate and test the models created is the accuracy, equation \ref{accuracy}, being $TP$ true positive, $TN$ true negative, $FP$ false positive and $FP$ false positive. This was the metric used because in this case we are only interested in knowing if the instruction was understood by the robot or not.

\begin{equation}
    Accuracy\; =\; \frac{TP\; +\; TN}{TP\; +\; TN \; + \; FP\; + \;FN} 
\label{accuracy}    
\end{equation}
\vspace{0.2cm}

\subsection{Test Datasets}

The various implementations were tested with two test sets, one related to the GPSR task and one with the FBM3. Both datasets are already in text to prevent the influence of speech recognition errors.
The test dataset for the GPSR was obtained using the generator provided by RoboCup and was annotated by hand. It contains $100$ instructions, being that some are composed by more than one command. 
For the FBM3, a dataset of the RoCKIn@Home in 2014\footnote{http://thewiki.rockinrobotchallenge.eu/index.php?title=Datasets}, competition that originated ERL, was used. It contains $180$ instructions, also with multiple commands. The annotations were already part of the set provided. Both sets are constituted by instructions that could be given from an human to a robot in a real context. 

\subsection{Word vectors comparison}

To compare the performance achieved using as input word embeddings created using the Word2vec and GloVe algorithms, and using one-hot vectors, both GPSR and FBM3 test sets were used.

In the tables \ref{test_word_embeddings_gpsr} and \ref{test_word_embeddings_fbm3}, the accuracy achieved for both the action detection and slot filling tasks, using the GPSR and FBM3 datasets, is presented, for all the word input vectors methods tested. The models used consisted in bidirectional LSTM with just one layer of $500$ cells for the action detection and slot filling.

\begin{table}[h]
  \renewcommand{\arraystretch}{} % more space between rows
  \centering
  \scalebox{1}{%
    \begin{tabular}[]{lccc}
      \toprule
      &   one-hot vectors &   GloVe    &   Word2vec\\
      \midrule
        action detection  &  0.826  &  0.934  &  0.863  \\[0.1cm]
        slot filling   &  0.817  & 0.895   &  0.871  \\
         \bottomrule
    \end{tabular}
  }%
    \vspace{0.1cm}
  \caption{Comparison of the performance when varying the type of word vectors, for GPSR}
  \label{test_word_embeddings_gpsr}%
\end{table}

\begin{table}[h]
  \renewcommand{\arraystretch}{} % more space between rows
  \centering
  \scalebox{1}{%
    \begin{tabular}[]{lccc}
      \toprule
      &  one-hot vectors  &  GloVe  &  Word2vec \\
      \midrule
        action detection   & 0.934  &  0.978  &  0.985   \\[0.1cm]
        slot filling   & 0.653  &  0.687  &  0.675  \\
    \bottomrule
    \end{tabular}
  }%
    \vspace{0.1cm}
  \caption{Comparison of the performance when varying the type of word vectors, for FBM3}
  \label{test_word_embeddings_fbm3}%
\end{table}

As expected the accuracy using one-hot vectors was the worst. This happened because the one-hot vectors do not capture the meaning of the words. This can be a problem when words do not appear in the training set but appear in the test set. 

Regarding the word embedding methods tested, GloVe and Word2vec skip-gram, the accuracy for both the action detection and the slot filling tasks is, in most tests, higher when using the feature vectors created by the GloVe algorithm. This was also an expected result, considering that in the comparison made in \cite{glove}, using a word analogy test, GloVe obtained a better performance. 

Consequently, in the following tests, the neural network inputs are GloVe word embeddings.

\subsection{Model improvements}

Having a working baseline model, we aimed at improving it. Some different neural network architectures, with varying sizes were tested. Also, a comparison between using implementation method $1$ and $2$, described in section \ref{sec:imple}, was made.

\subsubsection{Approaches comparison}

Firstly, a comparison of the two implementation methods was made.
To do this, a bidirectional LSTM neural network with $1$ layer of $500$ cells was tested for both test sets.

The results of the comparison for GPSR and FBM3 can be seen in table \ref{test_approaches}.

\begin{table}[htp]
  \renewcommand{\arraystretch}{} % more space between rows
  \centering

  {%
    \begin{tabular}[]{lcc}
      \toprule
                 &  Approach 1 \;\;\;\;  &  Approach 2 \;\;\;\;  \\
      \midrule
        GPSR Accuracy  & 0.895  & 0.874  \\[0.1cm]
        
        FBM3 Accuracy & 0.687  & 0.637  \\
        
     \bottomrule
    \end{tabular}
  }%
  \vspace{0.1cm}
  \caption{Comparison of the performance for the different approaches implemented}
  \label{test_approaches}%
\end{table}

As can be seen, the accuracy was slightly higher using approach $1$. This result was expected for GPSR and can be explained by the fact that the number of arguments that constitute the set, $6$, is not large, and some of them, such as $destination$ and $person$, can be present in a command of almost any action. However, once the arguments present in FBM3 dataset are more complex than in GPSR dataset, it was expected that approach $2$ would achieve a better result than method $1$. 
This result can be explained by the fact that in approach $2$, the selection of arguments depends on the action detection task, when errors occur while determining the action, the wrong model is used, increasing the chance of arguments being badly selected. 

Besides the better result, other reason to use approach $1$ is the fact that the computational cost when training is higher for approach $2$, since a network has to be trained for each of the actions while in approach $1$ only one network has to be trained. Also, when the system is being executed method $1$ is faster because the slot filling and action detection tasks run in parallel.
This way, in the following test the approach $1$ was used.

\subsubsection{Action detection}

Firstly, a comparison between the use of RNNs and LSTMs was made. The architecture used was similar, just replacing RNN cells by LSTM cells. The results for the GPSR  and FBM3 are showed in table \ref{test_commanddetectionrnn}.

\begin{table}[htp]
  \renewcommand{\arraystretch}{} % more space between rows
  \centering
  \scalebox{1}{%
    \begin{tabular}[]{lccccc}
      \toprule
      &   RNN  &   LSTM  \\
      & $1$ layer of $500$ cells & $1$ layer of $500$ cells \\
      \midrule
        GPSR Accuracy   & 0.869  &  0.931   \\[0.1cm]
        FBM3 Accuracy   & 0.894  &  0.978  \\
         \bottomrule
    \end{tabular}
  }%
    \vspace{0.1cm}
  \caption{Comparison of the performance in the action detection task when using RNNs and LSTMs}
  \label{test_commanddetectionrnn}%
\end{table}

As expected, the accuracy achieved when using LSTMs was higher. This can be caused by the existence of long-term dependencies that could not be memorized by the RNNs, since some of the commands in the datasets are quite extensive, reaching $15$ words.

Due to these results, different LSTM network architectures were evaluated.

\begin{table}[htp]
  \renewcommand{\arraystretch}{1} % more space between rows
  \centering
  {%
    \begin{tabular}{lc}
      \toprule
      & Accuracy\\
      \midrule
       LSTM - $1$ layer of $500$ cells   & 0.931 \\[0.1cm]
       DLSTM  - $2$ layers of $500$ cells  & 0.900 \\[0.1cm]
       BLSTM  - $1$ layer of $500$ cells  & 0.934 \\[0.1cm]
       DBLSTM  - $2$ layer of $250$ cells  & 0.923 \\[0.1cm]
       DBLSTM  - $2$ layers of $500$ cells  & 0.908 \\
         \bottomrule
    \end{tabular}
  }%
    \vspace{0.1cm}
  \caption{Comparison of the performance in the action detection task when varying the neural network architecture, for GPSR}
  \label{test_intentdetection_gpsr}%
\end{table}

Observing table \ref{test_intentdetection_gpsr}, it can be seen that there are no big differences in the accuracies obtained using the different types of LSTMs. Moreover, there is a small decrease in accuracy when the neural networks possess more than one layer. 
It can also be observed that the accuracy does not improve significantly when using bidirectional LSTMs. This result was expected, since the output of the command detection task consists in the final output of the sequence. 

Although the best result was obtained when using a neural network containing only $1$ layer of $500$ BLSTMS, the architecture chosen for the final model consists in a layer of $500$ LSTM cells, since the difference between the two accuracies is not substantial and the computational cost is lower for the simple LSTM, both when training and when executing the NLU system.

\begin{table}[htp]
  \renewcommand{\arraystretch}{} % more space between rows
  \centering
   {%
    \begin{tabular}{lc}
      \toprule
      & Accuracy\\
      \midrule
       LSTM  - $1$ layer of $500$ cells  & 0.978 \\[0.1cm]
       DLSTM  - $2$ layers of $500$ cells  & 0.969 \\[0.1cm]
       BLSTM  - $1$ layer of $500$ cells  & 0.978 \\[0.1cm]
       DBLSTM  - $2$ layer of $250$ cells  & 0.917 \\[0.1cm]
       DBLSTM - $2$ layers of $500$ cells  & 0.903 \\
         \bottomrule
    \end{tabular}
  }%
    \vspace{0.1cm}
  \caption{Comparison of the performance in the action detection task when varying the neural network architecture, for FBM3}
  \label{test_intentdetection_fbm3}%
\end{table}

Similarly as with GPSR dataset, using the FBM3 dataset it can be seen that the accuracy decreases when using multiple layers. The use of bidirectional LSTMs also decreased the performance.
The best results were obtained when a single layer of $500$ LSTM or BLSTM cells was used. 
Thus, also due to the BLSTM higher computational cost, the architecture chosen also consisted in a single layer of $500$ LSTM cells.

Comparing the performance in GPSR and in FBM3, the action detection results were higher in the FBM3. This can be a consequence of the FBM3 task having a smaller set of actions. 

\subsubsection{Slot filling}

Similarly as for the action detection task, the first comparison made when searching for the best neural network architecture to use was between RNNs and LSTMs.

\begin{table}[htp]
  \renewcommand{\arraystretch}{} % more space between rows
  \centering
   \scalebox{1}{%
    \begin{tabular}[]{lcc}
      \toprule
      &  RNN  &  LSTM \\
      & $1$ layer of $500$ cells & $1$ layer of $500$ cells \\
      \midrule
        GPSR Accuracy   & 0.836  &  0.873   \\[0.1cm]
        FBM3 Accuracy   &  0.584   &  0.637 \\
         \bottomrule
    \end{tabular}
  }%
    \vspace{0.1cm}
  \caption{Comparison of the performance in the slot filling task when using RNNs and LSTMs}
  \label{test_slot_filling_rnn}%
\end{table}

Observing table \ref{test_slot_filling_rnn}, it can be observed that the accuracy is better when using a single layer of $500$ LSTM cells than when a single layer of $500$ RNN cells is used. This happens for the same reason as in action detection, there are long-term dependencies that the RNNs are not able to capture.

Following, the performance when using different LSTM neural network architectures was analyzed.

\begin{table}[htp]
  \renewcommand{\arraystretch}{} % more space between rows
  \centering
   {%
    \begin{tabular}{lc}
      \toprule
      & Accuracy\\
      \midrule
       LSTM - $1$ layer of $500$ cells & 0.873 \\[0.1cm]
       DLSTM - $2$ layers of $500$ cells  & 0.922 \\[0.1cm]
       BLSTM - $1$ layer of $500$ cells  & 0.895 \\[0.1cm]
       DBLSTM - $2$ layer of $250$ cells  & 0.914 \\[0.1cm]
       DBLSTM - $2$ layers of $500$ cells  & 0.947 \\
         \bottomrule
    \end{tabular}
  }%
    \vspace{0.1cm}
  \caption{Comparison of the performance in the slot filling task when varying the neural network architecture, for GPSR}
  \label{test_slot_filling_gpsr}%
\end{table}

Differently than what happens in the action detection task, as can be seen in table \ref{test_slot_filling_gpsr}, the accuracy improves when more complex models, with more hidden layers, are used. This happens because the slot filling task corresponds to a more complex task, since it is a sequence tagging, an assignment of a categorical label to each member of the sequence, instead of a single classification. Also, the results obtained are better using bidirectional LSTMs, due to the fact that a classification of a word can depend on the following words instead of only on the previous words.

\begin{table}[htp]
  \renewcommand{\arraystretch}{} % more space between rows
  \centering
   {%
    \begin{tabular}{lc}
      \toprule
      & Accuracy\\
      \midrule
       LSTM - $1$ layer of $500$ cells  & 0.637 \\[0.1cm]
       DLSTM - $2$ layers of $500$ cells  & 0.648 \\[0.1cm]
       BLSTM - $1$ layer of $500$ cells  & 0.687 \\[0.1cm]
       DBLSTM - $2$ layer of $250$ cells  & 0.728 \\[0.1cm]
       DBLSTM - $2$ layers of $500$ cells  & 0.762 \\
         \bottomrule
    \end{tabular}
  }%
    \vspace{0.1cm}
\caption{Comparison of the performance in the slot filling task when varying the neural network architecture, for FBM3}
  \label{test_slot_filling_fbm3}%
\end{table}

Using the FBM3 dataset it was also determined, based on table \ref{test_slot_filling_fbm3}, that the results improve with the use of more than one layer and bidirectional LSTMs. The reasons that justify these results are the same as for the GPSR testset. 
Comparing the performance in the two tests, it can be observed that the accuracy is quite smaller for FBM3. This is due to the superior arguments complexity, i.e. in the GPSR test the arguments were composed by less words.

\subsection{ERL 2017}

For the 2017 ERL, a model, using this work, was created to compete in the FBM3 task. As the competition happened before the tests explained above were made, the model used was not the final one.
The system used was based on the second implementation approach, described in section \ref{sec:imple}.
Furthermore, for the action detection, a deep bidirectional LSTM with $2$ layers of $250$ cells was used, being the input word embeddings created using the Word2vec algorithm. 
For the slot filling, the architecture used in each of the models, one per action, was also deep bidirectional LSTM with $2$ layers of $250$ cells. The same word embeddings were used as input.

As far as we know, only other team, used an automatic natural language understanding system, LU4R.  As the list of verbs, names and objects were small and available to the teams, it was a possible to use different approaches, such as a simple parser.

 SocRob@home was placed in the third place, achieving a better performance than the team using LU4R. Although, the result would have probably been better if the model used had been the final one.

\section{Conclusion}

The purpose of this work was to develop a natural language understanding system that can be used by service robots, by detecting the action requested by a command given to a robot and its arguments. This was achieved by using deep neural networks, being RNNs and LSTMs evaluated for both steps.

We observed that LSTMs outperformed RNNs in the two tasks, action detection and slot filling. Considering LSTMs, bidirectional LSTMs had worse performances and the accuracy also decreased with the increase of complexity of the network, when detecting the action. On the other hand, for slot filling the results were superior when deeper bidirectional LSTMs were applied.

There are many possible extensions to this work, such as, using grounded information provided by the robots semantic map or developing a dialogue manager so the robot is able to continue the conversation with the human.

\section{Acknowledgements}

This work was supported by the FCT project [UID/EEA/50009/2013].

\bibliographystyle{IEEEtran}
\bibliography{IEEEexample}

\end{document}